\documentclass[runningheads]{llncs}

\usepackage{eccv}

\usepackage{eccvabbrv}

\usepackage{graphicx}
\usepackage{booktabs}
\usepackage{array}

\usepackage[accsupp]{axessibility}  
\usepackage[most]{tcolorbox}

\usepackage{xcolor}
\newcommand{\baseline}[1]{\textcolor{gray}{\textit{#1}}}

\usepackage{xspace}
\usepackage{hyperref}
\usepackage[capitalize]{cleveref}
\crefname{figure}{Fig.}{Figs}
\crefname{table}{Tab.}{Tabs}

\usepackage{caption}
\newcommand{\method}{\textsc{Molmo2Fish}\xspace}

\usepackage{hyperref}

\usepackage{orcidlink}

\begin{document}

\title{Teach a \method: Towards interactive fish tracking with natural language guidance} 

\titlerunning{Teach a \method}

\author{Kai van Brunt \and
Justin Kay\orcidlink{0009-0008-2344-5869} \and
Sara Beery\orcidlink{0000-0002-2544-1844}}

\authorrunning{K.~van Brunt et al.}

\institute{Massachusetts Institute of Technology, Cambridge MA 02139, USA\\
Correspondence to \email{kav@mit.edu}}

\maketitle

\begin{abstract}

Computer vision is increasingly used to automate recognition tasks in large ecological datasets, but more complex tasks such as multi-object tracking continue to pose challenges. As researchers seek to incorporate vision models in ecology workflows, various lines of research have explored how to make imperfect predictions useful through human-in-the-loop processes. We propose a new approach to working with imperfect tracking predictions through an interactive prediction correction workflow taking place as a conversation with a multimodal large language model, which we tailor to a sonar fish tracking dataset as an initial proof of concept. We investigate the performance of the tool, \method, across guided and unguided tasks, correcting its own predicted tracks and external tracks. We find that \method achieves high performance on fish tracking and track correction tasks, but there is still much room to improve on incorporating natural language guidance. The code and data are publicly available at \href{https://github.com/tidalove/molmo2fish}{github.com/tidalove/molmo2fish}.

\end{abstract}

\begin{figure}[t]
    \centering
    \includegraphics[width=.8\linewidth]{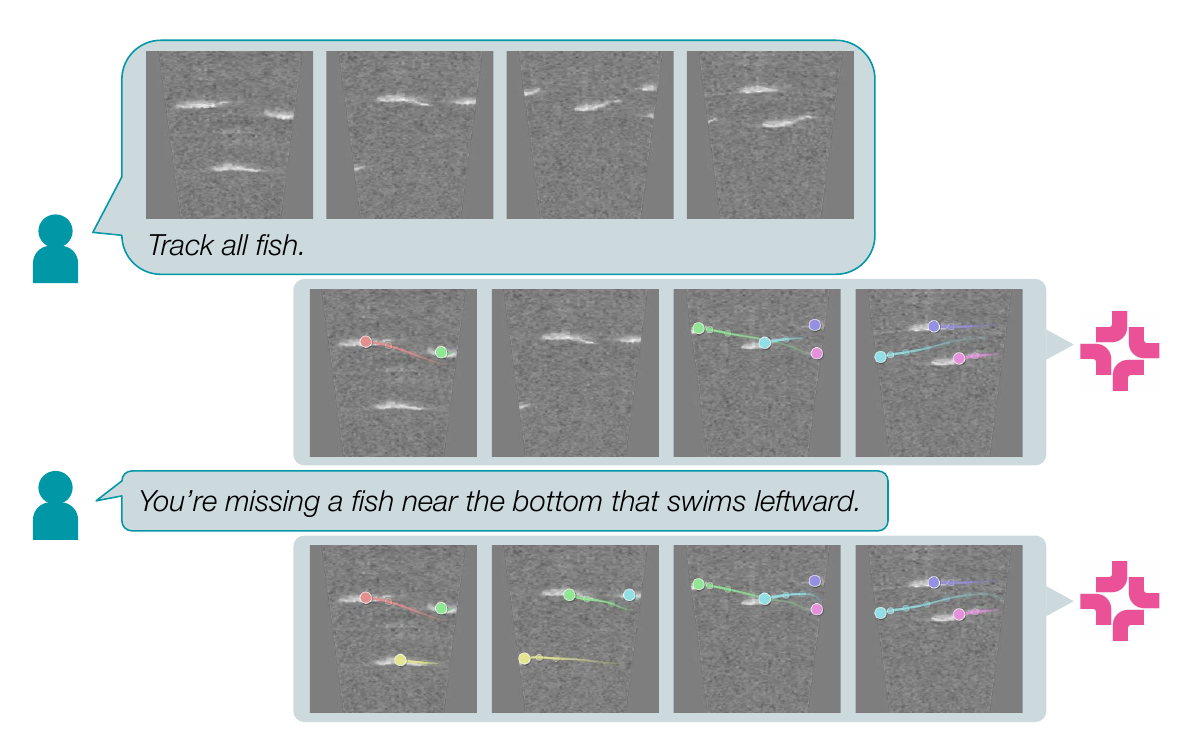}
    \vspace{-8pt}
    \caption{\method is an interactive tool for creating and editing multi-object tracking predictions. Users interact with a multimodal LLM that has been fine-tuned for fish tracking (top) and correcting tracks with natural language (bottom).}
    \label{fig:overall}
    \vspace{-12pt}
\end{figure}

\section{Introduction}
\label{sec:intro}

Computer vision is increasingly used in ecology and conservation to automate recognition tasks in large datasets, including species classification~\cite{hernandez2024pytorchwildlife,https://doi.org/10.1049/cvi2.12318}, object detection~\cite{beery2019efficient}, segmentation~\cite{Wasmuht2025-hk}, and multi-object tracking~\cite{Kay2022-bf,Dat2025-lb}. While detection and classification models have proven to be robust across many deployment scenarios in ecology~\cite{https://doi.org/10.1049/cvi2.12318,beery2019efficient}, more complex tasks such as multi-object tracking continue to pose challenges and require specialization to individual deployments~\cite{Kay2022-bf}. Improving the performance of tracking models could have significant impacts for ecology, enabling conservation use cases such as behavior analysis~\cite{Pedersen2020-eu}, population monitoring~\cite{Naik2024-di,Wasmuht2025-hk,atlas2023wild}, and real-time response to ecological incidents~\cite{Bondi2020-ja,Dat2025-lb}. 

In particular, we consider the use case of \textbf{counting migrating salmon in sonar video}~\cite{Kay2022-bf,atlas2023wild}. 
Sonar video cameras are deployed underwater in rivers to count the number of salmon migrating each year to spawn, informing the setting of fishing quotas and enabling monitoring of population response to conservation actions. 
Stakeholders have begun incorporating computer vision-based processing to reduce the time needed to process sonar video into salmon counts. 
This processing involves a specialized detection model and tracker that tracks fish as they enter and exit the field of view; counts are then determined according to the number of tracks.
While these specialized models perform well at river locations that appear in training data, their performance is brittle under environmental shift, causing prediction errors at new deployments and in changing environments~\cite{Kay2022-bf,kay2025align,kay2023unsupervised}.

Correcting erroneous tracking predictions involves collecting new ground truth annotations and re-training tracking models, presenting two barriers to practical use by fisheries stakeholders: \textbf{(1)}~multi-object tracking annotations are notoriously time-consuming to collect, requiring users to iteratively draw and adjust bounding boxes in potentially crowded scenes with complex movement~\cite{MAY2025103387,Mededovic2025-af}; and \textbf{(2)}~successfully re-training and validating models requires machine learning and data science expertise, which many fisheries stakeholders do not yet have. 

We propose a new approach to working with imperfect tracking predictions that addresses both of these problems simultaneously via an \textbf{interactive prediction correction workflow taking place as a conversation with a multimodal large language model (MLLM)}. 
MLLMs are trained to process visual and text inputs jointly through a large language model (LLM) architecture. Recent work has demonstrated these capabilities for multi-object tracking, in which an MLLM is prompted in natural language to track an object of interest in a video, and outputs text tokens specifying the locations of that object in each frame~\cite{molmo2openweightsdata}. 
We propose to extend these capabilities with a \textit{multi-turn} interaction in which the MLLM's initial predictions can be referenced and modified with additional natural language prompts (see Figure \ref{fig:overall}). 
This workflow addresses the problems above: 
instead of requiring manual labeling from scratch, users guide the model to produce better tracks; instead of requiring fine-tuning, the model updates its predictions in real-time conditioned on the user's correction prompts.

We evaluate the potential of our proposed workflow by training \method, a fine-tuned version of the open MLLM Molmo2~\cite{molmo2openweightsdata}, for tracking and iterative track correction in fisheries sonar video. 
Our experiments demonstrate first of all that MLLMs can indeed be specialized for fisheries sonar data through parameter-efficient fine-tuning~\cite{hu2021loralowrankadaptationlarge}, despite significant differences between sonar video and the typical MLLM training set.
We also demonstrate the potential to imbue MLLMs with multi-turn track correction capabilities that both improve overall tracking performance and target specific user-specified modifications.
However, there is still much room for improvement, and we discuss the potential for future work to improve upon our approach by leveraging more data and compute intensive training. 
To the best of our knowledge, our experiments are the first to empirically evaluate the capabilities of MLLMs in fisheries sonar, as well as the first to investigate and validate the proposed multi-turn track correction interaction. 
In summary, our contributions are:
\begin{enumerate}
    \item We propose a new approach to working with imperfect multi-object tracking predictions through a multi-turn interaction with a MLLM.
    \item We introduce \method, a custom MLLM that is trained to enable this workflow for the ecological use case of tracking salmon in sonar video.
    \item We empirically evaluate the performance of \method across both tracking and track correction tasks, providing insight into the current and future capabilities of using MLLMs for tracking and iterative prediction correction.
\end{enumerate}

\section{Related work}

\noindent \textbf{Multi-object tracking for ecology.}
Multi-object tracking (MOT) refers to the accurate localization and assignment to individual tracks of all objects of interest in a video. Traditionally, the most specialized and lightweight option has been the two-stage tracking-by-detection pipeline. An object detection model such as YOLO~\cite{cheng2024yoloworldrealtimeopenvocabularyobject}, Faster-RCNN~\cite{ren2016fasterrcnnrealtimeobject}, or SSD~\cite{Liu_2016} first predicts per-frame detections and classifications of all objects of interest. Next, an off-the-shelf tracker such as SORT~\cite{Bewley_2016} or ByteTrack~\cite{zhang2022bytetrackmultiobjecttrackingassociating}
links together detections across frames to form tracks. Such systems are often first-choice options for deployment because of their high speed and low demand for compute, but they are limited 
in their out-of-sample prediction ability, requiring additional supervised fine-tuning or domain adaptive training to address~\cite{kay2025align}. 

Ecological studies rely on detecting and tracking animals in videos to inform population monitoring~\cite{Kay2022-bf,Wasmuht2025-hk}, behavioral analysis in the wild~\cite{Naik2024-di,Gabeff2025-ck} and in situ~\cite{Pedersen2020-eu,Haalck2023-iy}, following animal populations in real time~\cite{Dat2025-lb}, preventing poaching~\cite{Bondi2020-ja}, and preventing accidents~\cite{Sun2024-zw}. Domain-relevant video object tracking datasets are collected and published to encourage development of deep learning methods that can infer such tracks from potentially huge amounts of video data efficiently and cost-effectively. We focus on the ecological use case of tracking fish in sonar video, building upon the Caltech Fish Counting (CFC) dataset~\cite{Kay2022-bf}. CFC is the largest open computer vision dataset for fisheries sonar, containing MOT annotations for 1,500 videos sourced from four different sonar deployments. We extend CFC with correction trajectories including natural language annotations to enable training \method for our proposed workflow.

\noindent \textbf{Multimodal LLMs.}
LLMs trained to interpret and answer questions about non-text modalities---image, audio, and video---have been growing in capability over the past few years. More recently, such multimodal LLMs (MLLMs) have been endowed with explicit video grounding capabilities: in addition to general video question answering, action recognition, and complex video understanding, increasingly MLLMs are trained to point at specific locations in a video as an auxiliary step in video reasoning, or as the final output to a pointing, tracking, or counting query. Gemini 3 Pro~\cite{gemini3pro2025} leads performance among proprietary models, while PerceptionLM~\cite{cho2025PerceptionLM,bolya2025PerceptionEncoder} and Qwen3-VL~\cite{bai2025qwen3vltechnicalreport} are examples of open models with such abilities. We elect to build on Molmo2~\cite{molmo2openweightsdata}, a fully open MLLM 
that achieves state-of-the-art performance on MOT tasks.
However, Molmo2---like most MLLMs---is trained predominantly on natural RGB footage. We show that it fails to generalize out of the box to fisheries sonar video, requiring fine-tuning.

\noindent \textbf{Working with imperfect predictions.}
As researchers seek to incorporate imperfect machine learning models in ecology workflows and data processing pipelines, an increasingly popular line of research in ML asks how to make imperfect predictions useful through human-in-the-loop processes. 
Methods for active \textit{inference} propose human-in-the-loop interactions that analyze or modify predictions directly. Existing approaches include statistical de-biasing~\cite{zrnic2024active}, transductive prediction~\cite{kurinchivendhan2026findingneedleshaystacktransductive}, and model selection~\cite{kay2025consensusdrivenactivemodelselection}. 

In computer vision, interactive video object segmentation
methods allow users to interact directly with predicted segmentations 
through point-and-click interactions that provide positive or negative feedback. 
Examples include SAM3~\cite{Carion2025-ux}, which propagates the correction in a learned fashion to other tracks and other frames, and
XMEM++\cite{Bekuzarov2023-sc} and EVA-VOS\cite{delatolas2024learning} that additionally suggest the next most useful frame for the user to annotate. \method differs by operating on point tracks rather than segmentations and by interacting through natural language on an entire video rather than direct user action on individual frames.


\section{\method}
\label{sec:method}

\method is a tool to interactively predict and correct tracks in video based on natural language prompts. Users participate in a multi-turn conversation with \method, prescribing specific corrections to tracks or nudging model predictions in the right direction by iteratively prompting the model with additional information (Figure \ref{fig:overall}). Initial predictions might be user-provided, generated by an external tracker, or generated by \method itself; \method is agnostic to the source of the tracks.

\subsection{Model details}

We build on Molmo2~\cite{molmo2openweightsdata}, a fully open MLLM trained at scale on a mixture of tasks which require spatial reasoning, video understanding, and precise video localization. In particular, Molmo2 can perform multi-object tracking of objects including people, animals, and automobiles, on a visually diverse set of videos. 

\begin{figure}[t]
    \centering\includegraphics[width=0.9\linewidth]{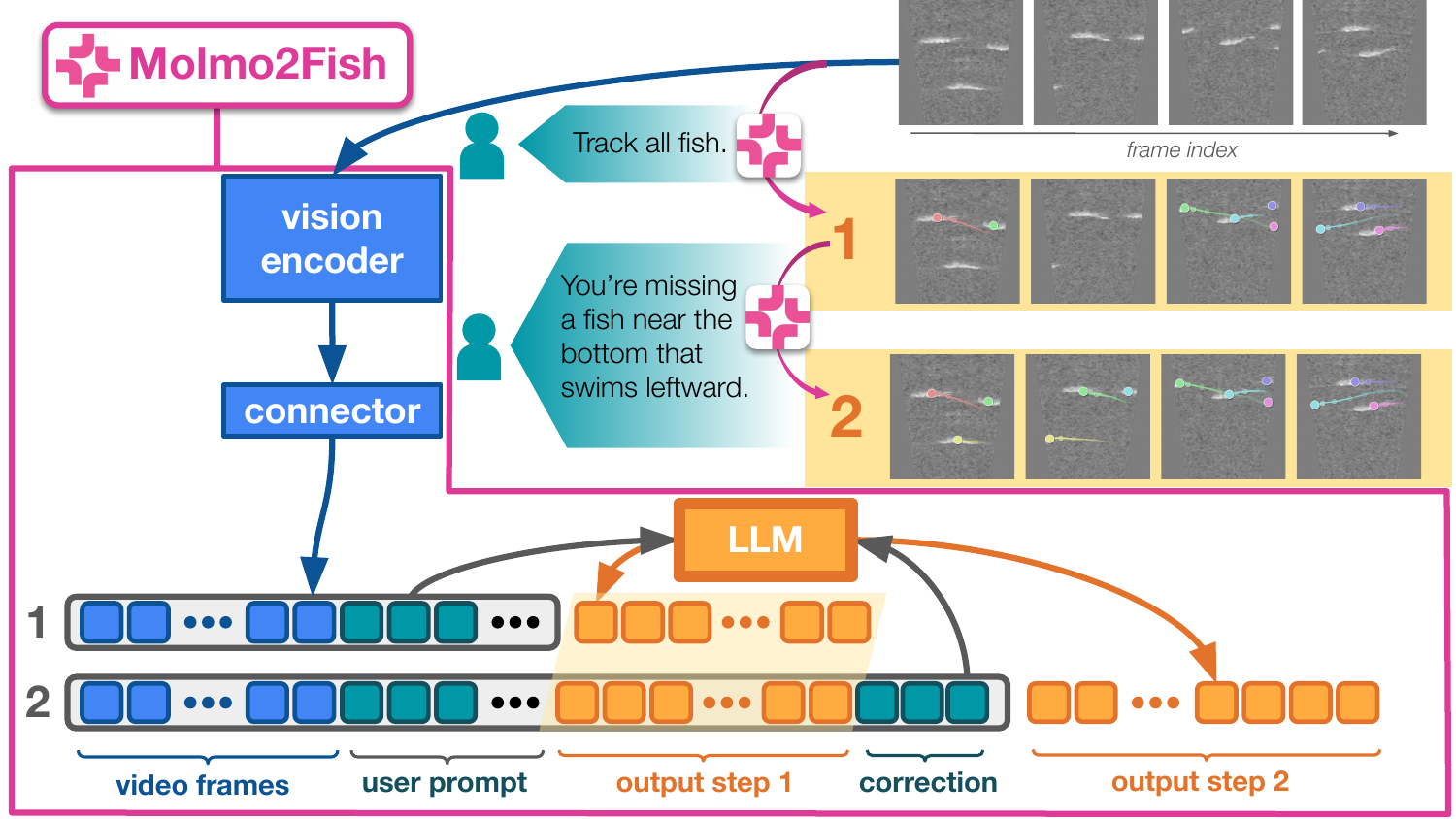}
    \vspace{-5pt}
    \caption{\textbf{\method architecture.} \method consists of the same three components as the Molmo2 MLLM we build upon: a video encoder, a LLM, and a ``connector'' that projects video representations into the LLM's representation space. Video frames and prompts are encoded and processed together, and outputs take the form of text token predictions. We extend the training protocol of the original Molmo2 to enable multi-turn corrections by constructing prompts as illustrated: first, the original video frames and a general prompt such as ``Track all fish''; then, a simulated text response from Molmo2 containing erroneous predictions; and finally a second textual correction prompt. 
    }
    \label{fig:tokens}
    \vspace{-12pt}
\end{figure}

Molmo2's architecture consists of three basic components, also illustrated in Figure \ref{fig:tokens}. The vision encoder (SigLIP 2) ingests video frames (potentially downsampled) spanning the length of the clip, encoding them into a latent visual feature space. The connector projects vision features to LLM embedding space so that they can be treated the same as language tokens by an LLM. The LLM (Qwen3) stacks vision tokens and user's text tokens in a single sequence and processes it.

Molmo2 outputs all responses as text. To specify the tracks it predicts for a video, it outputs a sequence of \texttt{\{t track\_id x\_coord y\_coord\}} strings, each denoting one instance's (\texttt{track\_id}) position (\texttt{x\_coord, y\_coord}) on one frame (at time \texttt{t} in seconds). If an object of a particular \texttt{track\_id} is not present at time \texttt{t} then Molmo2 neglects to output it in that step of the sequence. The sequence order is fixed by sorting all ground truth points by \texttt{(t, x\_coord, y\_coord)}, \textit{i.e.} the earliest points in the sequence are those which come earliest temporally, followed by those with a lower x coordinate, then those with a lower y coordinate.

\subsection{Instruction tuning for track correction}

Molmo2's video understanding and grounding capabilities form a strong starting point for our tool, but it was trained only for one-step, non-interactive tasks---it has no notion of referring back to a previous step to inform its next output. Hence, to build \method we fine-tune Molmo2 concurrently on two objectives: \textbf{(1) a \textit{pure tracking} task} of the same format as tracking tasks in the Molmo2 training corpus, in order to improve domain-specific tracking performance, and \textbf{(2) an interactive \textit{track correction} task}, which Molmo2 has not seen before, that enables interactive track correction as illustrated in Figure~\ref{fig:overall}. In Section \ref{sec:data} we describe the composition of data used to train \method.

The text content of each training example for the \textit{pure tracking} task takes the same form as tracking tasks in Molmo2's training set. In our training, these examples look like:

\begin{tcolorbox}[
    enhanced,
    colback=gray!10,
    frame hidden,
    arc=3mm,
    boxsep=1pt
]
\texttt{User:} $\langle$\texttt{video}$\rangle$ \texttt{Track all fish.}

\texttt{Assistant:}
\end{tcolorbox}

The text content of each training example for the \textit{track correction} task takes the form of an artificial multi-turn conversation between the user and LLM which fits entirely in the LLM's context window. A track correction example looks like:

\begin{tcolorbox}[
    enhanced,
    colback=gray!10,
    frame hidden,
    arc=3mm,
    boxsep=1pt
]
\texttt{User:} $\langle$\texttt{video}$\rangle$ \texttt{Track all fish.}

\texttt{Assistant:} $\langle$\texttt{corrupted\_point\_sequence}$\rangle$ 

\texttt{User:} $\langle$\texttt{correction\_prompt}$\rangle$ 

\texttt{Assistant:}
\end{tcolorbox}

We refer to the sequence \texttt{(i: \{prompt, tracks\}) for i in num\_steps} constituting a track correction example as a \textit{trajectory}. In practice, due to context window length limitations, in our experiments we train and evaluate on only the single-step case (one correction step).

The model is trained with cross entropy loss between its output logits and the tokens corresponding to the ground truth point sequence string. In a \textit{pure tracking} task, this corresponds to the model's immediate prediction (following ``Track all fish'') of all tracks. In a \textit{track correction} task, this corresponds only to the model's response following the user's correction prompt.

We limit our work to finetuning on a single dataset (described in Section \ref{sec:cfc}) for faster training, easier correction prompt generation, and as a proof of concept that can motivate future work to develop a more general model.

\section{Dataset}
\label{sec:data}

We design \method with a specific ecological use case: tracking fish in sonar video. To do so, we extend the Caltech Fish Counting (CFC) dataset~\cite{Kay2022-bf}, the largest open dataset of fisheries sonar video, to enable the interactive workflow described in Section \ref{sec:method}. Here, we describe CFC in more detail and how we adapted it for \method, with more detailed dataset statistics in \cref{tab:dataset}.

\begin{table}[h]
\caption{\textbf{Dataset tasks.} Overview of all tasks used in the dataset, for both training and evaluation.}
\vspace{-8pt}
\label{tab:subsets}
\setlength{\tabcolsep}{9pt}
\centering
\scriptsize{
\begin{tabular}{lp{6cm}l}
\toprule
\footnotesize \textbf{Name} & \footnotesize \textbf{Description} & \footnotesize \textbf{Splits} \\
\midrule
Pure tracking & Generic ``track all fish'' task, target is all points on all fish on all frames & train, val \\
\addlinespace
Guided tracking & ``Track X fish'', where X is some descriptor: ``track the last fish'', ``track the top three fish'', etc; target is points only on those identified fish & train, val \\
\midrule
Synthetic track corrections & Tracking corrections generated synthetically & train, val \\
\addlinespace
Targeted track corrections & Tracking corrections fixing only target mistakes & train, val \\
\addlinespace
Real track corrections & Tracking corrections generated from real model predictions: & \\
\addlinespace
\quad YOLO+SORT & Tracking corrections with pre-correction prediction from YOLO+SORT predictions & val only \\
\addlinespace
\quad Molmo-low & Tracking corrections with pre-correction prediction from step110 \method checkpoint & train, val \\
\addlinespace
\quad Molmo-high & Tracking corrections with pre-correction prediction from step300 \method checkpoint & train, val \\
\bottomrule
\end{tabular}}
\end{table}

\subsection{Caltech Fish Counting} 
\label{sec:cfc}

CFC consists of sonar videos of salmon annotated with bounding boxes and tracks. Tracks can be used to derive upstream and downstream counts of traveling salmon, which are used for estimation of salmon escapement statistics.


The original dataset contains data from four different sonar deployments: the \textbf{Kenai} left bank, Kenai \textbf{right bank}, \textbf{Elwha}, and \textbf{Nushagak} rivers. These locations are visually distinct and together span a range of tracking conditions in fisheries sonar data, from the Elwha river's small, sparse fish to the Nushagak's dense clusters of large fish.

We construct training and validation sets from the union of these river locations. We randomly select 25\% of video clips for validation and train on the remainder for Elwha, Nushagak, and Rightbank. We use the original paper's split of Kenai data.
This training/validation split, which incorporates data evenly from across these locations, is different from the original CFC training/validation split, which was designed to evaluate OOD generalization by training on a single location and holding out every other location. 


\subsection{Trajectory generation pipeline}\label{sec:corruption}

\begin{figure}[t]
    \centering
    \includegraphics[width=.8\linewidth]{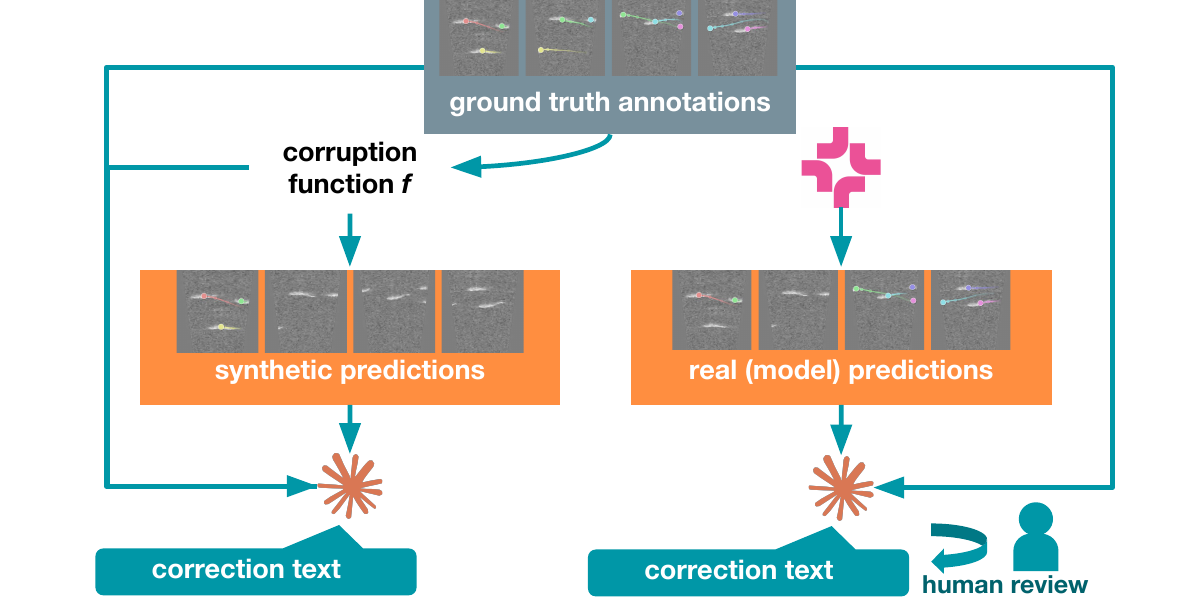}
    \vspace{-8pt}
    \caption{From a video and its corresponding ground truth annotations, we construct three sets of predictions to use as the ``pre-correction'' step during training and validation: ``synthetic'' which applies a corruption function to ground truth, and ``Molmo-low'' and ``Molmo-high'' which come from a model's tracking predictions on the clip.}
    \label{fig:pipeline}
    \vspace{-12pt}
\end{figure}

We build a set of \textit{synthetic correction trajectories} with CFC videos and annotations as source material. We generate multiple correction trajectories per video, each consisting of: \textbf{(1)} an initial set of imperfect tracks, \textbf{(2)} a natural language correction prompt that addresses one or more tracking errors from (1), and \textbf{(3)} the intended, post-correction set of tracks.

\noindent \textbf{Classifying tracking mistakes.} We construct artificially corrupted correction trajectories, inspired by common failure modes in multi-object tracking. 
Previous work on evaluating multi-object trackers classifies tracking errors into five types: false negatives, false positives, fragmentations, mergers, and deviations~\cite{luiten2020IJCV}. We build upon these corruption primitives to construct a set of possible corruptions enumerated in Table \ref{tab:corruptions}, including more complex patterns that are constructed from the sequential application of multiple primitives.

\begin{table}[t]
\caption{All possible corruptions to apply to ground truth tracks (possibly in combination) to generate the trajectories in synthetic correction dataset.}
\vspace{-8pt}
\label{tab:corruptions}
\centering
\setlength{\tabcolsep}{9pt}
\scriptsize{
\begin{tabular}{lp{\dimexpr\linewidth-4cm\relax}}
\toprule
Corruption & Function \\
\quad \texttt{add} & Adds a spurious track \\
\quad \texttt{delete} & Removes one track \\
\quad \texttt{fragment} & Splits a track into two, assigning one track ID before a chosen frame and another after \\
\quad \texttt{merge} & Picks two tracks to assign to the same track ID \\
\quad \texttt{deviate} & Applies a spatial displacement \\
\quad \texttt{partial\_delete} & Removes a portion of a track \\
\quad \texttt{duplicate} & Adds a spurious track close to a track \\
\quad \texttt{drift} & Adds a directional shift to a track \\
\quad \texttt{jitter} & Adds random noise to a track \\
\quad \texttt{stall} & After some frame, track stays still \\
\quad \texttt{id\_switch} & Two tracks are covered by one (one track moves from one fish to another) \\
\quad \texttt{split\_gap} & One track is split into two, with some intermediate missing coverage \\
\quad \texttt{swap} & Two tracks have their IDs swapped at a certain frame \\
\quad \texttt{light\_subsample} & Reduces up to 50\% of annotated points per track \\
\quad \texttt{delete\_most} & Deletes all but 1--2 ground truth tracks per video, leaving them untouched \\
\quad \texttt{subsample} & Reduces annotated points per track, up to retaining only 2 points per track \\
\bottomrule
\end{tabular}}
\vspace{-12pt}
\end{table}

\noindent \textbf{Synthetic corrections.} From each clip's ground truth tracks we apply corruptions programmatically to arrive at the initial, step-1 tracks. The model tries to correct \textit{all} corruptions and recover the ground truth tracks in step 2.

\noindent \textbf{Targeted corrections.}
We also include a set of \textit{targeted} corrections in the set of synthetic corrections. For this set we write prompts asking to correct only the $n$ most recent corruptions of total $N$ corruptions applied to the ground truth tracks. The post-correction tracks the model learns to generate are thus the ground truth tracks with only the first $N-n$ corruptions applied. We also compute HOTA against that flawed ground truth. To perfectly execute a targeted correction, the model \textit{must} pay attention to the user prompt: it cannot just rely on learning to correct as well as possible in response to every video.

\noindent \textbf{Prompt variety.}
A human user desiring a specific correction to an incorrect track prediction may issue a variety of prompts to \method, differing in style and information content. We thus include prompts with a range of information content.

\textit{No-info} prompts read as ``Fix all mistakes in these tracks, if any exist.'' The correct output may be a duplicate of the pre-correction tracks. \textit{Wrong-only} prompts read as ``Fix all mistakes in these tracks.'' \textit{Vague} prompts ($\leq20$ words) specify what corrections to make but don't include timestamps or exact counts of dense fish clusters, and sometimes don't refer precisely to track IDs. \textit{Full} prompts ($\leq40$ words) fully specify what corrections to make, including timestamps, exact counts of dense fish clusters, and track IDs. In practice, we find a difference of $<1\%$ across the board in final HOTA between \textit{no-info} and \textit{wrong-only} evaluations, and between \textit{vague} and \textit{full} evaluations, so for concision we omit all results on \textit{wrong-only} and \textit{vague} prompts throughout the paper.

\begin{figure}[h]
    \centering
    \includegraphics[width=.9\linewidth]{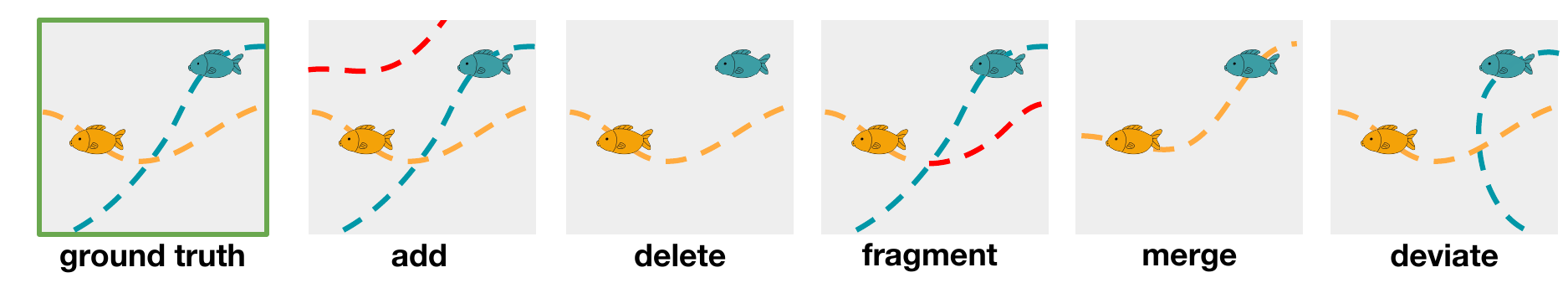}
    \vspace{-8pt}
    \caption{An illustration of the \texttt{add}, \texttt{delete}, \texttt{fragment}, \texttt{merge}, and \texttt{deviate} corruptions, with line color corresponding to predicted track ID.}
    \label{fig:corruptions}
    \vspace{-12pt}
\end{figure}

\noindent \textbf{Writing correction prompts.}
Multiple sets of corruptions are applied to a video's ground truth tracks to construct trajectories of varying correction difficulty and complexity. To generate the correction prompt for a trajectory, an LLM (Opus 4.5) receives a record of which corruptions were applied in what order and the sorted track IDs computed before and after each corruption.

\noindent \textbf{Correcting Molmo's predictions.}
To bolster \method's self-correction ability, we additionally include a set of correction trajectories whose pre-correction steps are Molmo2's predictions on the video. We fine-tune a version of Molmo2 only on the tracking task and not on the correction task and save its predictions on the entire CFC dataset to use as the pre-correction state. The step110 checkpoint, with lower performance, produces the ``Molmo-low'' set; the step300 checkpoint, with higher performance, produces the ``Molmo-high'' set.

We provision an Opus 4.5 agent
to generate natural language correction texts for each case. We manually reviewed the subset of the resulting prompts corresponding to the most structurally complex videos to edit prompts for accuracy. See Appendix \ref{sec:generation} for more detail on all data generation steps.

\section{Experimental setup}

\noindent \textbf{Training configuration.}
For the LLM to finish generating a correction, its context window must fit the video, initial prompt, initial prediction, user correction prompt, and final prediction. For CFC clips which are too long to fit, we slice them into parts so they fit fully in the context (see the \textit{Sliced clips} row of Table \ref{tab:dataset}), in addition to downsampling frames by a factor of 3 when fed into the model so the video can span a greater temporal extent.

We perform rank 64 LoRA~\cite{hu2021loralowrankadaptationlarge} finetuning on all components with a learning rate of $0.001$. We LoRA finetune as opposed to fully finetuning because of computational limitations. We train all models for 300 steps with a batch size of 64 and a maximum sequence length of 16,384 tokens. The model is trained with cross-entropy loss on its final response only (\textit{i.e.} the output step it predicts, either step 1 for the pure tracking task or step 2 for the track correction task).

\noindent \textbf{Evaluation metrics.}
We evaluate multi-object tracking performance using Higher Order Tracking Accuracy (HOTA)~\cite{luiten2020IJCV} which measures both detection accuracy and long-range association accuracy over all tracks in a video (see Appendix \cref{sec:hota} for more details). To accommodate the differences between Molmo2's tracking outputs (points) and CFC's annotation format (bounding boxes), we modify HOTA 
such that a true positive always refers to a point landing in its ground truth box; we thus do not average over any localization thresholds as in the original implementation.




\noindent \textbf{Evaluation protocol.}
To understand \method's correction ability we employ the following protocol. From multiple sets of initial predictions produced by various models run on the entire CFC validation set of videos, we generate correction prompts (of all four \textit{full}, \textit{vague}, \textit{wrong-only}, and \textit{no-info} varieties), as described in Section \ref{sec:data}, and run a single correction step on each video. We then compare the step-1 (pre-correction) HOTA averaged over the whole validation set and the step-2 (post-correction) average HOTA to measure the change in tracking performance resulting from the correction step.

We evaluate \method on three sets of initial (step-1) predictions, also described in \cref{tab:subsets}: 
\noindent \textbf{(1) YOLO+SORT predictions.}
The CFC authors built a lightweight two-stage object detection and tracking pipeline using the object detector YOLO~\cite{cheng2024yoloworldrealtimeopenvocabularyobject} and off-the-shelf tracker SORT~\cite{Bewley_2016}. 
We follow their procedure but train on our training set as described in \cref{sec:cfc}.
We also evaluate corrections to step-1 tracks generated by \method in response to the prompt ``Track all fish'', \textbf{(2) Molmo-low predictions} from an early-checkpoint \method model trained only on tracking and \textbf{(3) Molmo-high predictions} from the final tracking-only checkpoint.

\section{Results}

\begin{figure}[t]
    \centering    \includegraphics[width=.9\linewidth]{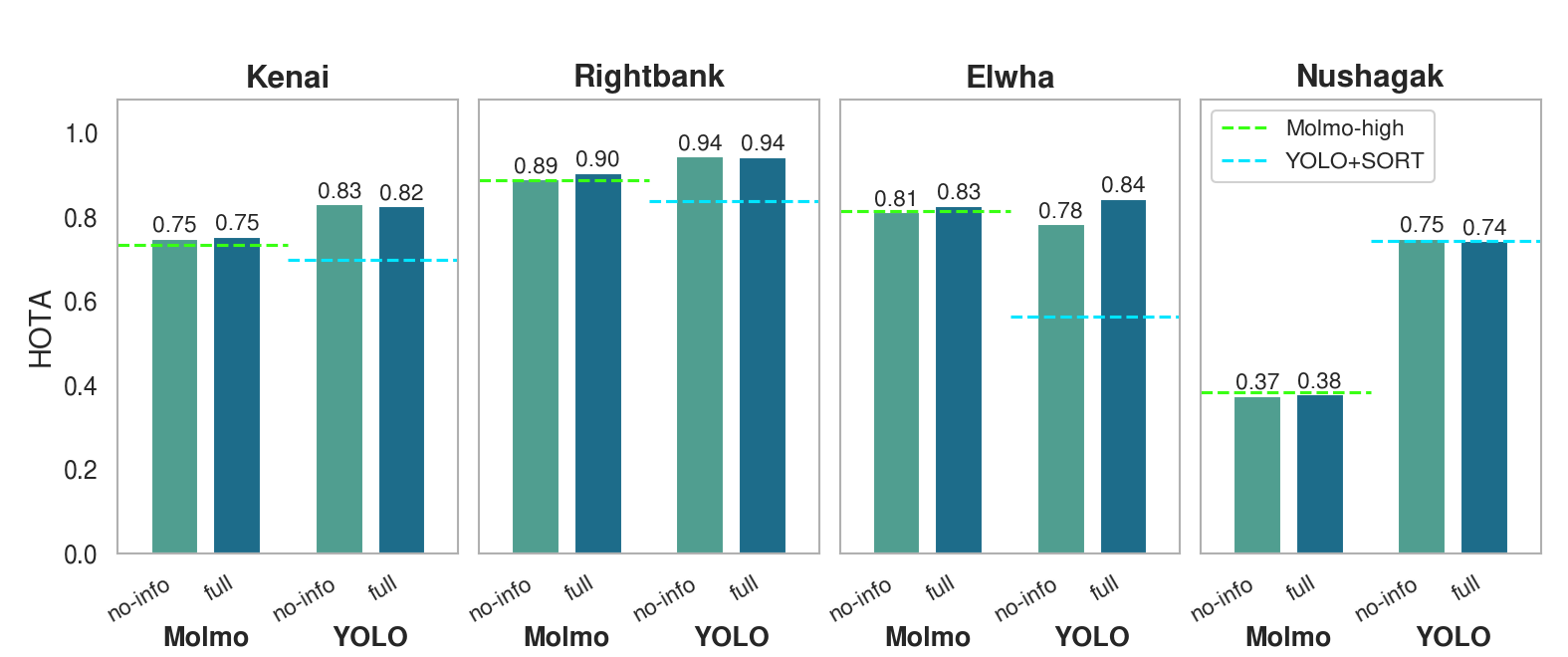}
    \caption{HOTA of pre-correction (dashed lines) and post-correction (bars) predictions on each river. Pre-correction predictions are generated by Molmo and YOLO+SORT respectively; post-correction predictions are generated by \method from either the Molmo-high or YOLO+SORT baseline.}
    \label{fig:corrections}
    \vspace{-12pt}
\end{figure}

\textbf{Correcting model-generated predictions.} When evaluated with the above single-step correction protocol, \method's post-correction prediction improves from the YOLO+SORT prediction baseline on three out of four test locations (see \cref{fig:corrections}, left three panels). This is true regardless of whether prompts provide specific corruption-correction information or not---both \textit{no-info} and \textit{full} prompts lead to tracking improvements.
On the other hand, when starting from Molmo-high (\textit{i.e.}, the strongest Molmo-generated prediction set), improvements are less consistent and generally more modest. This indicates that while \method succeeds at correcting predictions that differ from its own, it struggles to further improve its own best predictions beyond their initial accuracy even with targeted prompting.
We also observe that \method struggles both pre- and post-correction on videos that contain many fish, e.g. at the Nushagak location where tracks are dense. See Appendix \ref{sec:examples} for examples.

\noindent
\textbf{Ablating fine-tuning strategies.} We ablate parameters fine-tuned in \cref{tab:trainablation} and data included in training in \cref{tab:dataablation}. In our parameter fine-tuning ablation, all versions of the model are trained on the same data mixture of pure tracking and synthetic, targeted, and Molmo-low corrections. Fine-tuning more components of the model improves performance significantly across all tasks except on Molmo-high corrections, which all versions of the model perform similarly at. Some fine-tuning is necessary to make Molmo2 useful for the CFC data and on the track correction task: its out-of-the-box performance (the leftmost column in \cref{tab:trainablation}) is low on both evaluations. In our training data ablation, all models have LLM, connector, and ViT LoRA fine-tuned, for the same number of steps. Again, some fine-tuning on the track correction task is necessary to make the model useful for that task: the model trained on pure tracking only (the leftmost column in \cref{tab:dataablation}) performs poorly on the synthetic correction task even though it leads at pure tracking. Only the models trained on Molmo-low (rightmost two columns in \cref{tab:dataablation}) meaningfully correct \textit{both} the synthetic and Molmo-low pre-correction baselines. Molmo-high, which starts from a baseline closer to ground truth, is more difficult for all models to correct, even the model in the rightmost column of \cref{tab:dataablation} which has seen Molmo-high trajectories during training.

\begin{table}[t]
\caption{\textbf{Fine-tuning ablation.} HOTA on each validation set for Molmo2 out-of-the-box (no fine-tuning) and Molmo2 trained by fine-tuning the LLM only, LLM and connector only, and all components. For all \textit{correction} tasks, the pre-correction HOTA of the step-1, corrupted tracks is listed in gray next to the task name, and the reported score is HOTA on the entire validation set after a single correction step on each video.}
\vspace{-8pt}
\label{tab:trainablation}
\centering
\setlength{\tabcolsep}{6pt}
\scriptsize{
\begin{tabular}{lcccc}
\toprule
 & Molmo2 & LLM & LLM+Connector & LLM+Connector+ViT \\
\midrule
Pure Tracking & 0.049 & 0.608 & 0.656 & 0.737 \\
\midrule
Synthetic Corrections
& \baseline{0.484} & 
\baseline{0.484} & \baseline{0.484} & \baseline{0.484} \\
\quad No-info & --- & 0.757 & 0.769 & 0.799 \\
\quad Full & 0.164 & 0.767 & 0.780 & 0.817 \\
\midrule
Targeted Corrections
&  & 
\baseline{0.843} & \baseline{0.843} & \baseline{0.843} \\
\quad No-info & --- & 0.623 & 0.632 & 0.640 \\
\quad Full & --- & 0.895 & 0.899 & 0.900 \\
\midrule
Molmo-low Corrections
&  & 
\baseline{0.482} &
\baseline{0.482} & \baseline{0.482} \\
\quad No-info & --- & 0.514 & 0.528 & 0.615 \\
\quad Full & --- & 0.548 & 0.580 & 0.669 \\
\midrule
Molmo-high Corrections
&  & \baseline{0.737} & 
\baseline{0.737} &
\baseline{0.737} \\
\quad No-info & --- & 0.739 & 0.741 & 0.737 \\
\quad Full & --- & 0.740 & 0.749 & 0.747 \\
\bottomrule
\end{tabular}}
\vspace{-12pt}
\end{table}

\begin{table}[h]
\caption{\textbf{Training dataset ablation.} HOTA after LoRA fine-tuning all model components for 300 steps on each dataset mixture, for a model trained only on the tracking task; only on tracking and synthetic corrections; on tracking, synthetic corrections, and the Molmo-low set; and on all training sets. (We include an extra rightmost column reflecting a training mixture containing all sets showing that performance can be further improved by training for more steps, but opted to compare all other models at 300 steps.) For all \textit{correction} tasks, the pre-correction HOTA of the step-1, corrupted tracks is listed in gray next to the task name, and the reported score is HOTA on the entire validation set after a single correction step on each video. Only the models that saw Molmo-low data during training make appreciable corrections to Molmo-low trajectories. Despite training on Molmo-high, the rightmost columns' models still fail to make significant corrections to Molmo-high trajectories.}
\vspace{-8pt}
\centering
\setlength{\tabcolsep}{3pt}
\scriptsize{
\begin{tabular}{lcccc|c}
\toprule
 & Pure tracking & + synthetic & + Molmo-low & + Molmo-high & \baseline{+120 steps} \\
\midrule
Pure Tracking & 0.780 & 0.763 & 0.737 & 0.729 & 0.793 \\
\midrule
Synthetic Corrections
& \baseline{0.484} & \baseline{0.484} & \baseline{0.484} & \baseline{0.484} & \baseline{0.484} \\
\quad No-info & --- & 0.820 & 0.799 & 0.766 & 0.803 \\
\quad Full & 0.415 & 0.833 & 0.817 & 0.811 & 0.848\\
\midrule
Molmo-low Corrections
& & \baseline{0.482} & \baseline{0.482} & \baseline{0.482} & \baseline{0.482} \\
\quad No-info & --- & 0.473 & 0.615 & 0.567 & 0.576 \\
\quad Full & --- & 0.500 & 0.669 & 0.634 & 0.694 \\
\midrule
Molmo-high Corrections
& & \baseline{0.737} & \baseline{0.737} & \baseline{0.737} & \baseline{0.737} \\
\quad No-info & --- & 0.740 & 0.737 & 0.744 & 0.743 \\
\quad Full & --- & 0.755 & 0.747 & 0.754 & 0.767 \\
\bottomrule
\end{tabular}}
\label{tab:dataablation}
\vspace{-12pt}
\end{table}

\begin{figure}
    \centering
    \includegraphics[width=1\linewidth]{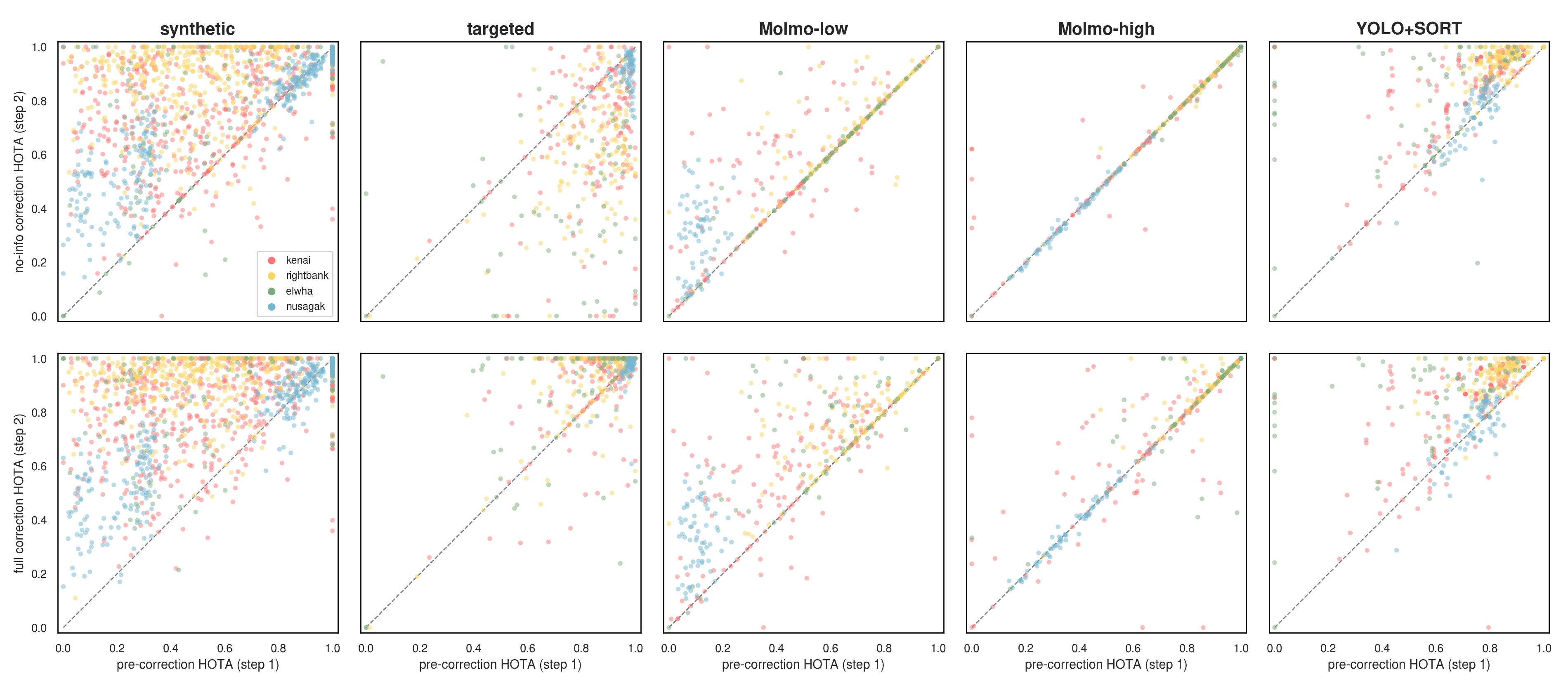}
    \vspace{-12pt}
    \caption{HOTA before and after correction, with no-info prompts (top row) and full prompts (bottom row), for each subset of the track correction evaluation data. Each dot represents a single trajectory. For the targeted track correction task (where HOTA of 1.0 corresponds to the partially-corrupted ground truth as described in \cref{sec:corruption}), a generic ``Fix all tracks'' prompt results in a large aggregate drop in HOTA (top row), while \method makes the targeted correction successfully when the prompt specifies what to fix (bottom row). In other tasks, the change between no-info and full corrections is subtler.}
    \label{fig:hota-before-after}
    \vspace{-12pt}
\end{figure}

\noindent\textbf{\method attends to the user's correction prompt.}
``Targeted'' corrections are evaluated against the \textit{partially corrected} corrupted tracks and not against the fully correct ground truth, as described in \cref{sec:corruption}. The correction prompt must specify which corrections to make, and \method must listen, to make such corrections successfully. As expected, and visible in \cref{fig:hota-before-after}, \method consistently improves over the pre-correction baseline when provided with a fully detailed prompt, and consistently degrades the tracks relative to the target when provided with a generic prompt of the form ``Fix all tracks'', which leads it to fix mistakes that are left in the targeted ground truth (``Targeted Corrections'' row of \cref{tab:trainablation}).

\noindent \textbf{More information doesn't consistently help.}
The rightmost column of \cref{tab:trainablation} shows \method improves significantly over the pre-correction baseline for synthetic corrections, but incorporating prompts with full detail only modestly improves its predictions. Fully detailed prompts significantly improve \method's performance on Molmo-low, but in that case, the corrected tracks still lag behind the model's one-shot tracking performance. On Molmo-high, \method barely improves over the pre-correction baseline and is helped only slightly by detailed prompts. Similarly, in \cref{fig:corrections}, when correcting from the YOLO+SORT baseline (rightmost group of bars in each chart), \method performs only marginally better given detailed prompts (``full'' bar) vs. without any information (``no-info'' bar).

\section{Discussion and conclusions}

We successfully adapted Molmo2 to tracking in a new visual domain and to a new mode of interaction: generating better tracks based on flawed initial predictions. We found that LoRA finetuning is sufficient to introduce these capabilities to Molmo2, which is unable to complete these tasks zero-shot, and that training on our data increases tracking accuracy on the CFC dataset from just 5\% to 74\%.
Further, building on a model that initially has no notion of referring to its previous output, 
we showed it is possible to add track correction capabilities using our data generation and fine-tuning framework.
\method is able to perform targeted corrections, and is also a strong general-purpose track correction model, often improving predictions even without specific guidance. 

However, there are limitations and room for improvement. 
\method is sensitive to distribution shift between the set of corruptions and correction prompts seen during training and at test-time.
Further, while it is promising that \method's initial tracking performance is so high, we observed that additional natural language feedback often does not help the model to substantially improve upon its own initial performance. Simultaneously, it's clear that \method pays attention to the user's correction prompt: it successfully makes targeted corrections when provided with a prompt describing the specific mistakes to fix and fails without a prompt, as it should.

To interpret \method's behavior on non-``targeted'' track corrections, we can consider the correction trajectories as existing across a difficulty spectrum. For the easiest trajectories, whether \method is provided a detailed prompt or not, it is able to make the desired correction. For the hardest trajectories, whether \method is provided a detailed prompt or not, it is unable to make the desired correction. Only for trajectories occupying an intermediate range on this difficulty spectrum does \method's behavior change depending on the presence of a detailed prompt. Empirically, it seems that the prevalence of examples occupying this intermediate range in the dataset is quite low, so that in our evaluation setup, the improvement to aggregate HOTA over the whole dataset is just a few percent at most. The project of training a more capable track correction model would endeavor to understand the dynamics within and/or increase the ceiling on this intermediate range: a model which could more intelligently use language and video in tandem may be able to leverage language priors to improve on those more difficult predictions.

On the other hand, for the targeted track correction task, regardless of the difficulty of correcting the initial prediction, \method \textit{must} attend to the user prompt to understand which corrections to apply or not, so in aggregate, a much larger improvement is seen between the no-info and full evaluations.

In conclusion, \method is a model that allows for interactive track correction in the form of a conversation with an MLLM. We investigated what works, what doesn't, and how future development might bring us closer to a model that can help users interact with and modify tracking predictions on video in a flexible manner, reducing the effort needed to develop and test ecological workflows on tracks in video.

\section*{Acknowledgements}
Thanks to Michael Hobley, Suzanne Stathatos, Madison Van Horn, Sevan Brodjian, and Erik Young for their insights and feedback throughout.

%
%
\bibliographystyle{splncs04}
\bibliography{main}

\clearpage

\appendix

\section{\method prediction examples}
\label{sec:examples}

Here we provide examples of model behavior on both the \textit{tracking} task (given just the frames of a video, output tracks for all fish) and the \textit{correction} task (fill in the last step of a correction trajectory). We describe \method's strengths and weaknesses, including direct comparisons to the YOLO+SORT traditional two-stage object detection and tracking method that we compare against throughout this work. 

\subsection{Tracking examples}

\begin{figure}
    \centering
    \includegraphics[width=\linewidth]{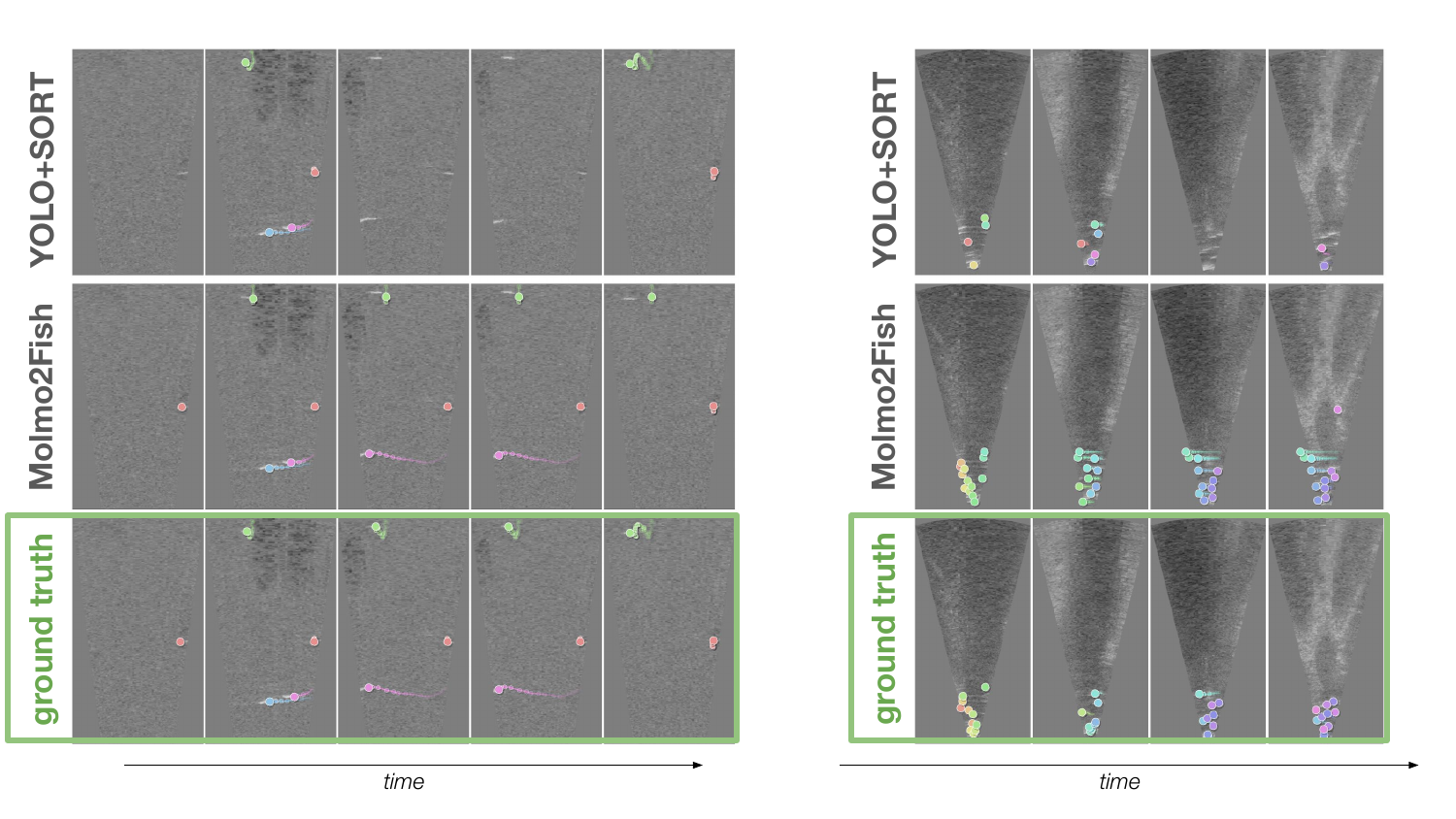}
    \caption{Examples of \method's tracking predictions (middle row) compared against YOLO+SORT's tracking predictions (top row) on two clips, from the Rightbank (left group) and Nushagak (right group) locations. Left: \method successfully carries some tracks through the whole video that YOLO+SORT drops on some frames, but \method's predicted track on the topmost fish stalls and stops following it, while YOLO+SORT's prediction doesn't. Right: \method successfully localizes fish on more frames for more of the video but there are so many fish that it ``loses track'' and begins to extrapolate tracks horizontally rather than ending them.}
    \label{fig:trackingexamples}
\end{figure}

We show examples of \method's tracking predictions next to YOLO+SORT's predictions in \cref{fig:trackingexamples} and share general observations on qualitative trends in its predictions compared to those output by YOLO+SORT.

\method's baseline performance is much higher on the Elwha river than YOLO+SORT. The fish in the Elwha river appear small and are often occluded by noise and sonar reflections. On the Elwha, an understanding of the motion of the fish over the entire clip is helpful for localizing it on any given frame. While YOLO tends to miss detections on individual frames where the fish is less visible, leading SORT to fragment a single fish into multiple tracks, \method learns what a coherent fish track looks like over the course of a video and maintains the same track when appropriate. In addition, the Elwha is relatively sparse---not many fish swim through the sonar camera's field of view at the same time---avoiding a weakness of \method, which struggles with keeping track of many fish.

On the other hand, \method performs much worse than YOLO+SORT on the Nushagak, where fish appear large and easy to localize, but many fish swim through at the same time. In fact Molmo2 was not trained on scenes with above 40 tracks, and the Nushagak clips routinely bump up against that upper limit, with 30 to 40 fish swimming through in a single clip. While the YOLO+SORT algorithm is equally robust regardless of clip length or total number of tracks, \method gets confused once the track count becomes large, forgetting to drop tracks once they exit the frame or extrapolating tracks horizontally.

\subsection{Correction examples}

\begin{figure}
    \centering
    \includegraphics[width=.9\linewidth]{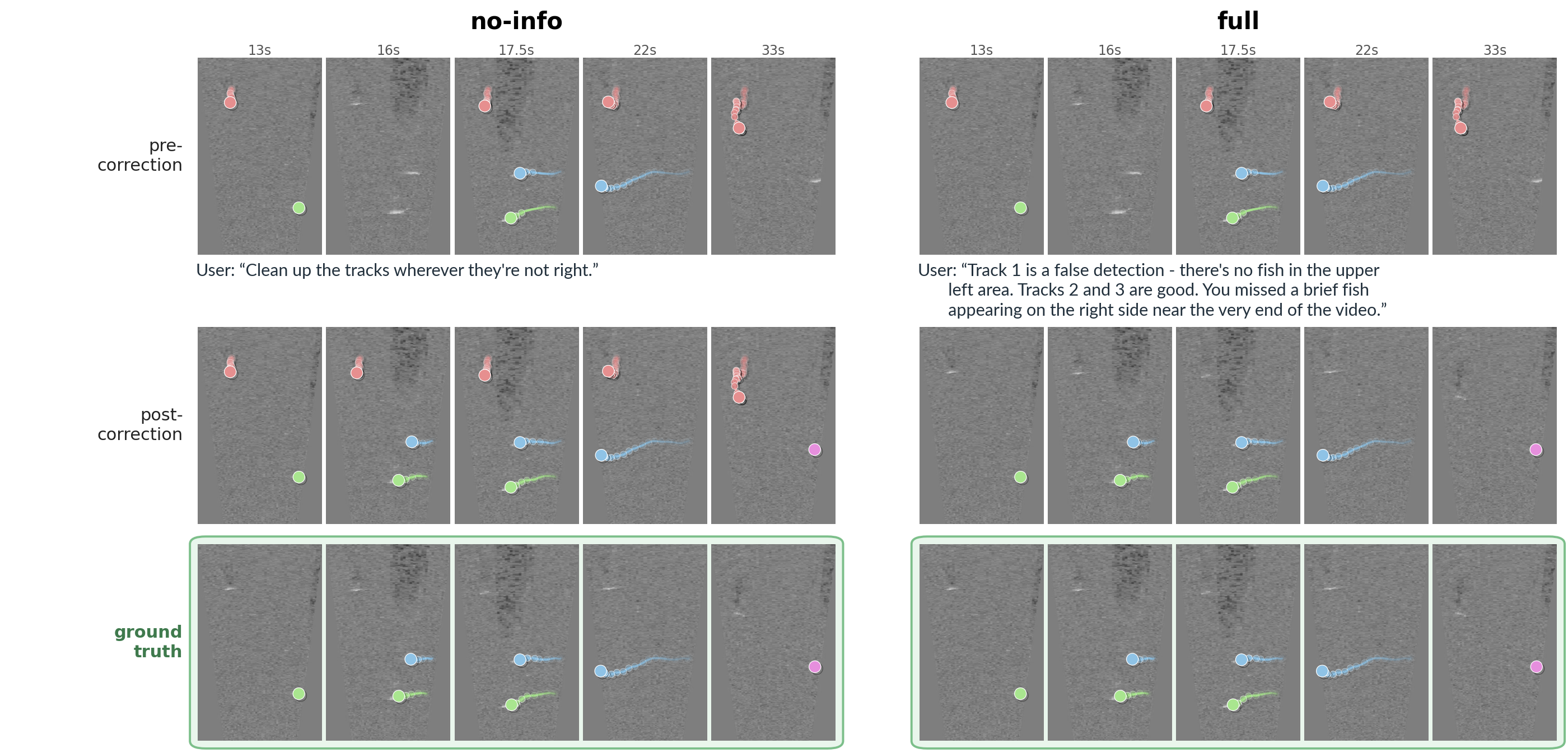}
    \caption{Example of \method making a better correction (second row) from the same pre-correction step (top row) when provided with a detailed prompt (right) vs when provided with a generic prompt (left). The ground truth tracks are displayed on the same frame sequence in the bottom row for comparison. Here, \method recognizes the upper left blob as a fish unless explicitly told by the user that it is a false detection.}
    \label{fig:good-correction-1}
\end{figure}
\begin{figure}
    \centering
    \includegraphics[width=.9\linewidth]{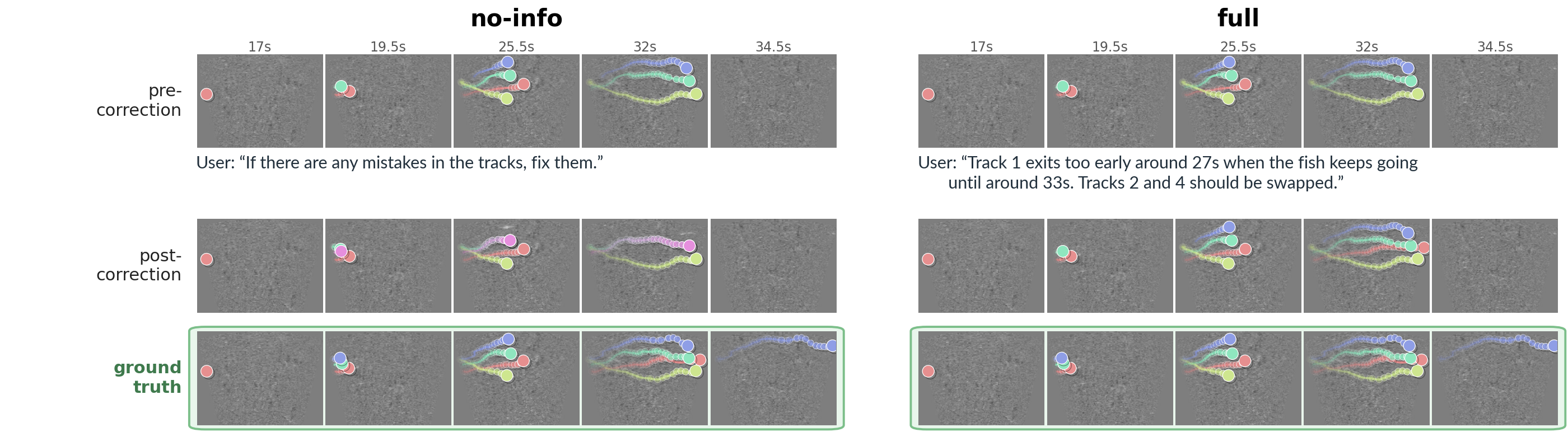}
    \caption{Similar to \cref{fig:good-correction-1}, an example where \method makes a better correction, as measured by HOTA, when provided with a detailed prompt vs without. \method thinks the mistake is that the top fish (blue) is a false detection: actually the mistake pointed out by the user is that the first track (red) must be present longer.}
    \label{fig:good-correction-2}
\end{figure}

In \cref{fig:good-correction-1} and \cref{fig:good-correction-2} we show correction examples where \method makes a better correction from step zero (top row of each group) when given a fully detailed prompt (right column) than when given a generic prompt (left column). In all cases, the target tracks are the ground truth tracks, shown in the bottom row of each group. When the prompt helps, it's often by helping \method make judgment calls more aligned with the human annotator's judgment. 

\begin{figure}
    \centering
    \includegraphics[width=.9\linewidth]{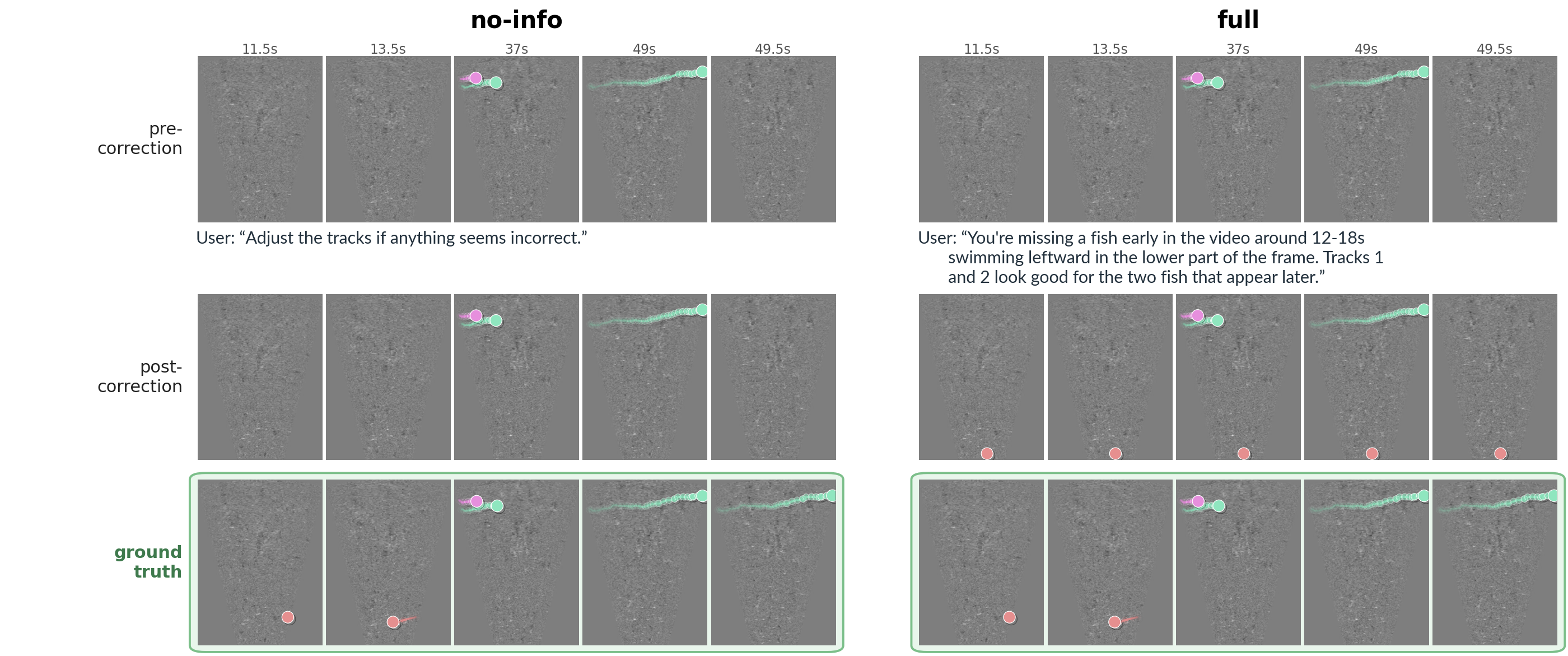}
    \caption{Example of \method making a \textit{worse} correction when provided with a detailed prompt vs without. Since \method can't recognize the presence of the difficult-to-localize fish that appears early in the video, it merely refrains from making a correction at all when prompted to ``Adjust the tracks if anything seems incorrect'', whereas it puts a point down randomly (and incorrectly) when told that an additional fish does exist in that region.}
    \label{fig:bad-correction-1}
\end{figure}

\begin{figure}
    \centering
    \includegraphics[width=.9\linewidth]{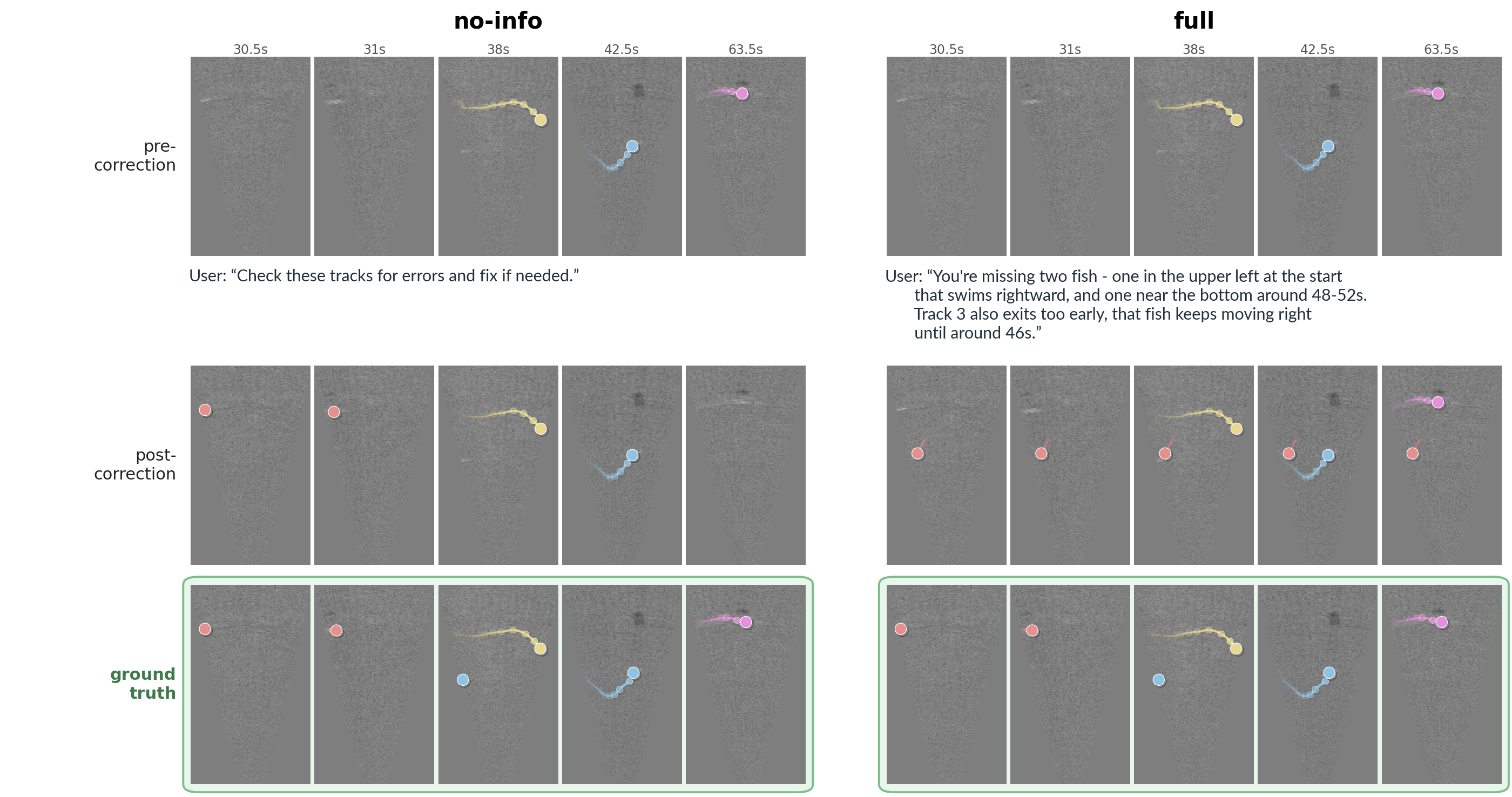}
    \caption{Example of \method making a \textit{worse} correction when provided with a detailed prompt vs without. Here \method is diverted from its initially correct prediction of the first fish's track by the user prompt.}
    \label{fig:bad-correction-2}
\end{figure}

We also show examples in \cref{fig:bad-correction-1} and \cref{fig:bad-correction-2} of the fully detailed prompt \textit{hurting} \method's correction performance. These cases are usually explained by the prompt neglecting to mention some flaw in the initial tracks that \method fixes when told to ``fix any mistakes'' but does not fix when specific corrections are prompted that don't include that initial flaw. For example, when correcting from the YOLO+SORT baseline, it often happens that the ``full'' prompt does not capture the intermittency of the YOLO+SORT tracks.

\clearpage

\section{Additional experiments}

\subsection{Generalization challenges}

We investigate the ability of \method to generalize to corruption types not seen during training in \cref{tab:ood}.
We hold out a set of corruption types (\texttt{split\_gap}, \texttt{swap}, and \texttt{duplicate}) during training on only synthetic trajectories and 
measure \method's 
performance on a set of test clips including those held-out corruptions. 
Holding out \texttt{swap} means that during training the model \textit{never sees} correction trajectories that undo \texttt{swap}, regardless of whether \texttt{swap} is explicitly named in the prompt. 
We call trajectories containing these corruptions the ``OOD set'' and trajectories not containing these corruptions the ``ID set''.

We see that distribution shift in corruptions/corrections causes poor performance. 
The two models perform similarly when evaluated on the ID set. But evaluated on the OOD set (top row group of \cref{tab:ood}), with or without language guidance, the model trained only on the ID set (left column) lags behind the model that saw all corruptions during training (right column).

\begin{table}[h]
\caption{HOTA on evaluation of models trained on pure tracking and either a) ID set, only synthetic trajectories not including held-out corruptions or b) all synthetic trajectories. Both models perform similarly evaluated on the ID set, which contains corruptions both saw during training. A large performance gap exists between the two models when evaluated on the OOD set, which is not recovered by natural language guidance.}
\vspace{-8pt}
\setlength{\tabcolsep}{6pt}
\label{tab:ood}
\centering
\small{
\begin{tabular}{llcc}
\toprule
& & Trained on ID only & Trained on all \\
\midrule
OOD set & \baseline{Pre-correction baseline} & \baseline{0.628} & \baseline{0.628} \\
             & No-info    & 0.810 & 0.856 \\
             & Full       & 0.824 & 0.882 \\
\midrule
ID set & \baseline{Pre-correction baseline} & \baseline{0.432} & \baseline{0.432} \\
             & No-info    & 0.805 & 0.806 \\
                     & Full       & 0.823 & 0.815 \\
\bottomrule
\end{tabular}}
\vspace{-12pt}
\end{table}

\clearpage

\section{Dataset details}

\begin{table}[h]
\caption{\textbf{Full dataset statistics.} Split by location, for the train and validation sets. ``Clips'' refers to number of clips in original CFC dataset: we randomly allocate clips to the train or validation set before slicing to fit in \method's context window.}
\label{tab:dataset}
\setlength{\tabcolsep}{9pt}
\centering
\small{
\begin{tabular}{llrrrr}
\toprule
& & Kenai & Rightbank & Elwha & Nushagak \\
\midrule
\multicolumn{6}{l}{\textbf{Train}} \\
\midrule
\textit{Clips} & & 482 & 512 & 175 & 55 \\
\textit{Sliced clips} & & 858  & 844  & 273 & 154 \\
\textit{Examples} & Pure tracking    & 858  & 844  & 273 & 154 \\
         & Guided tracking & 2048 & 2130 & 492 & 449 \\
         & Synthetic corrections       & 2104 & 2116 & 404 & 602 \\
         & Molmo-low corrections           & 856  & 844  & 273 & 142 \\
         & Molmo-high corrections           & 858  & 843  & 273 & 154 \\
         & Targeted corrections      & 788  & 646  & 144 &  72 \\
\midrule
\multicolumn{6}{l}{\textbf{Validation}} \\
\midrule
\textit{Clips} & & 64 & 144 & 48 & 17 \\
\textit{Sliced clips} & & 159 & 144 &  69 &  70 \\
\textit{Examples} & Pure tracking    & 159 & 144 &  69 &  70 \\
         & Guided tracking & 539 & 567 & 153 & 240 \\
         & Synthetic corrections       & 435 & 372 &  93 & 280 \\
         & Molmo-low corrections            & 159 & 144 &  69 &  67 \\
         & Molmo-high corrections           & 159  & 144  & 69 &  69 \\
         & Targeted corrections       & 158 & 144 &  68 &  70 \\
\bottomrule
\end{tabular}}
\end{table}

\newpage

\section{Data generation process}
\label{sec:generation}

Below we describe the process of generating correction prompts for both the real and synthetic correction datasets.
We use the open-source software \texttt{ramure}~\cite{ramure2026} to provision LLMs, to expose tools to them, and to collect their responses in a reproducible, programmatic fashion. 

\subsection{Real corrections}

To generate the pre-correction step for the trajectories making up the ``real'' correction dataset, we collect predictions on the training set and validation set of a version of Molmo2 fine-tuned only on fish tracking at each of step 110 and step 300 of its training. We refer to these sets as `Molmo-low' and `Molmo-high' respectively throughout the paper. Corrections are written for each trajectory by an LLM which has access to:

\begin{itemize}
    \item the incorrect tracks
    \item an initial attempt (deterministic Hungarian matching) at a track assignment between each predicted track and a ground truth track
    \item the ground truth tracks
    \item a set of tools (function calls) that can help it understand high-level track features
\end{itemize}

The LLM's system prompt is shown below.

\begin{tcolorbox}[
    enhanced,
    breakable,
    colback=gray!10,
    frame hidden,
    arc=3mm,
    boxsep=1pt,
    fontupper=\small
]

You review fish tracking predictions and write brief, casual correction feedback.

Your job has two parts:

\textbf{Part 1: Review Track Assignments}

The deterministic matching algorithm has made initial assignments (predicted tracks → ground truth tracks). Review these — they may be wrong. Common issues to look for:

\begin{itemize}
    \item Two predicted tracks may be FRAGMENTS of one fish (temporally sequential, spatially continuous → should be merged)
    \item A track marked "spurious/false positive" may actually be a fragment of a nearby matched track
    \item A track marked as matching one GT may actually correspond to a different GT
\end{itemize}

Look at timing, spatial position, and motion direction to judge whether assignments make sense.

\textbf{Part 2: Write Correction Text}

Write a single correction message (1-3 sentences max, casual tone).

Rules:
\begin{itemize}
    \item Be brief and casual (1--3 sentences max)
    \item Reference prediction track numbers (Track 1, Track 2, etc.)
    \item Use vague time references: ``around 35s'', ``near the end'', ``early on'', ``at the start''
    \item Use spatial descriptions for missing fish: ``between tracks 3 and 4'', ``in the center'', ``near the top''
    \item For videos with many tracks or many errors: be MORE vague and holistic (e.g., ``your tracks aren't exiting at the right time'')
    \item For videos with few tracks and specific errors: be MORE specific (e.g., ``Track 2 stays on the frame much longer than it should'')
    \item Acknowledge correct tracks briefly when some are wrong (e.g., ``Track 3 is good'')
    \item Never use pixel coordinates or technical jargon
    \item Sound natural, like a human giving quick feedback
    \item Don't start with ``I'' or ``The model'' --- address the tracker directly with ``you/your'' or just describe what's wrong
\end{itemize}

Error patterns you may see:
\begin{itemize}
    \item fragmented: model splits one fish into multiple sequential tracks (should be merged)
    \item stall: track freezes in place while fish keeps moving
    \item drift: track gradually moves away from the actual fish position
    \item offset: track follows the fish's motion but is consistently shifted away from it
    \item early\_exit: track ends too early (fish is still there)
    \item late\_exit: track stays too long (fish has already left)
    \item late\_entry: track starts too late (fish was already there)
    \item diverge: track suddenly jumps away from the fish
    \item spurious: a predicted track with no real fish there
    \item missing: a real fish that wasn't tracked at all
\end{itemize}

Examples of good correction text:
\begin{itemize}
    \item ``Track 2 stays on the frame much longer than it should. The fish exits around 35s''
    \item ``Track 1 follows the fish correctly but it should exit around 40s. Track 2 follows the fish correctly at first but loses it around 37s, that fish keeps drifting upwards to the left and then downwards to the left.''
    \item ``All the tracks stall at some point. Tracks 1, 2, and 3 even start off stalled and don't follow their respective fish as they're moving.''
    \item ``you missed a fish at the beginning in the bottom part of the video that exits out the left. tracks 1 and 2 should be merged. Tracks 3 and 4 should be merged, and are missing some coverage in the middle of the track''
    \item ``you're missing a fish swimming between tracks 3 and 4''
    \item ``your tracks aren't exiting at the right time - the fish are continuously exiting the frame out the left''
    \item ``you're missing a fish in the center at the start of the clip. Both tracks 1 and 2 start drifting upwards of their actual fish and are present for too long. Track 3 is good.''
    \item ``tracks 1 and 2 are drifting above the actual locations of the fish''
    \item ``there are only 2 fish, not 3, and they start swimming downwards at around 37s, diverging from the predicted tracks''
\end{itemize}

\textbf{Output format}

\begin{verbatim}
<confidence>high or low</confidence>
<reasoning>1-2 sentences on why you're confident or not 
--- what's ambiguous?</reasoning>
<correction>your correction text here</correction>
\end{verbatim}

Mark confidence as LOW when ANY of these apply:
\begin{itemize}
    \item There are tracks marked ``spurious'' that overlap temporally/spatially with other tracks (could be fragments)
    \item Multiple short-lived tracks (< 10 frames) exist --- they might be fragments of longer fish
    \item The number of predicted tracks doesn't match the number of GT tracks AND the discrepancy isn't easily explained by a single missing/spurious track
    \item A track's error could reasonably be described in multiple different ways
    \item You had to choose between two plausible interpretations of the track structure
\end{itemize}

Mark confidence as HIGH only when ALL of these apply:
\begin{itemize}
    \item Track assignments are clearly one-to-one with no ambiguity
    \item Errors are simple and unambiguous (clear stall, clear late\_exit, obvious single missing fish)
    \item No short-lived tracks that could plausibly be fragments
    \item Your correction would be the same regardless of how you interpret the ambiguous parts
\end{itemize}

\end{tcolorbox}

We manually review all trajectories marked low confidence for accuracy. We observe empirically that most errors in captioning originate from model assignments between predicted and ground truth tracks that don't align with how a human would make the assignments. As a simple example, consider a clip with just one fish and two predicted tracks. Depending on the tracks' proximity to the ground truth fish track, the LLM may claim that track 2 is spurious while a human may claim that track 2 and track 1 are part of the same track, or vice versa. These are subjective judgments which we address by writing prompts ourselves, but the ambiguity

\subsection{Synthetic corrections}

Each video gets four synthetic trajectories, with each trajectory resulting from a number of stacked corruptions that ranges depending on the number of fish in the video clip. For clips with two or fewer fish, all trajectories contain up to four corruptions at most. For clips with three or more fish, one trajectory contains 1-2 corruptions; one contains 3-4 corruptions; another 4-6 corruptions; another 7+ corruptions. ``One corruption'' is one application of one of the functions listed in \cref{tab:corruptions}.

A log is constructed which contains the order of corruptions applied, all pre-corruption (ground truth) track IDs, and all post-corruption (seen by the model) track IDs. This is important so that the LLM correctly refers to the track IDs which are seen by the model and assigned deterministically.

Given this log and the same set of tools as in the real data generation pipeline in case the LLM wants to look at the actual tracks (e.g. to inform some phrasing), the LLM is asked to rephrase the log into a natural language correction prompt, following some guidelines. The full system prompt is reproduced below.

\begin{tcolorbox}[
    enhanced,
    breakable,
    colback=gray!10,
    frame hidden,
    arc=3mm,
    boxsep=1pt,
    fontupper=\small
]

You turn a structured fish-tracking correction LOG into ONE natural-language correction a human reviewer would give. The log is an ORDERED list of every issue with the model's tracks, joined by commas (and `then' before the last). SUMMARISE the WHOLE list into a single coherent correction that covers EVERY item — do not drop any, do not invent any. Merge naturally where several items concern the same fish, and keep a sensible order.

Notation: most items name tracks as [step0\_id, step1\_id] — step0 is the track id the model currently shows (what the reviewer points at), step1 is the id it should become; None means the track is absent from that step. When two items share the same step1 id, they are the same fish.

Issue vocabulary:

\begin{itemize}
    \item add track [X, None]: a spurious track on no real fish — should be removed.

    \item delete track [None, X]: a real fish the model missed — should be added.

    \item stall (mid): a track frozen in place mid-way while the fish keeps moving — the frozen stretch should follow the fish (endpoints are unchanged; this is NOT an early/late issue).

    \item 'frozen in place past frame N ... trim its end back to frame N' / 'frozen in place before frame N ... trim its start forward to frame N': an EDGE stall — the track is stuck on a frozen box and OVERSHOOTS where the fish really is, so it must be SHORTENED (trimmed) to frame N. Do NOT say it stops early or should be extended — it runs too long/starts too early.

    \item 'starts late ... extend its start back to frame N' / 'stops early ... extend its end to frame N': frames were dropped from one end so the track is TOO SHORT and must be LENGTHENED (extended) to frame N. This is the OPPOSITE of an edge stall.

    \item jitter / drift: a track drifting off the fish.

    \item split / near-duplicate / split-with-gap: TWO tracks that are really one fish (should be merged into one track).

    \item split-into-N: N tracks (3 or 4) that are all really ONE fish, broken into consecutive fragments (should all be merged into one track).

    \item `missing some of its points' / `make it continuous': a track that exists but has gaps — some frames were dropped; it should be filled in to be continuous (NOT added from scratch, NOT merged).

    \item `trying to cover both tracks i and j': one track that drifts from one fish onto another (an id switch) and should be two separate tracks.

    \item merge: two fish collapsed into one track (should be split apart).

    \item swap: two tracks' ids are swapped.

    \item `sparse' / `fill in the rest of the points': a track that exists but is missing many of its points throughout — it should be densified, NOT added from scratch.

    \item `pick up the rest of the fish tracks': most fish are untracked; track the remaining fish (sometimes naming the frames at which they pass by, and an exact count for a dense cluster).
    
\end{itemize}

STYLE — write like a clear human note:

\begin{itemize}
    \item HARD CAP: 40 words maximum.
    \item Frame numbers from the log are FRAME NUMBERS — you MAY keep them (a later stage converts frames to seconds). Do NOT convert them yourself or invent times.
    \item If the log states an exact count for a dense cluster of fish, you MAY cite it.
    \item Never mention pixels or raw coordinates; say `upward', `to the left', etc.
    \item You have geometry tools (track positions, distances, frame snapshots). They are OPTIONAL — use them only if you want to refer to a fish by its position relative to other tracks (e.g. `the leftmost track', `the fish below track 2').
\end{itemize}

Call submit\_correction exactly once with the correction text. Do not say anything else.

\end{tcolorbox}

We also inject prompts randomly (50-50) with one or the other of the following style note since we observe anecdotally that otherwise the LLM is biased heavily towards using descriptive phrasing.

\begin{tcolorbox}[
    enhanced,
    breakable,
    colback=gray!10,
    frame hidden,
    arc=3mm,
    boxsep=1pt,
    fontupper=\small
]

\begin{itemize}
    \item STYLE FOR THIS ONE: PRESCRIPTIVE — phrase it as an instruction/action (e.g. `add the track on the fish at the top', `merge tracks 2 and 3', `extend track 4 to the end').
    \item STYLE FOR THIS ONE: DESCRIPTIVE — phrase it as an observation of what's wrong (e.g. `you missed a fish at the top', `tracks 2 and 3 are the same fish', `track 4 stops too early').
\end{itemize}

\end{tcolorbox}

\subsection{Targeted corrections}

Targeted corrections are generated through the same pipeline as synthetic corrections except that the step-1 target is not the video ground truth (GT) but a corrupted version of the video GT. The LLM captioner additionally just receives the last $n$ corruptions to write a prompt for.

\section{HOTA}
\label{sec:hota}

To calculate HOTA, first we arrive at a bijective matching of predicted detections to ground truth detections. True positives (TPs), false positives (FPs), and false negatives (FNs) are determined as usual from this matching. 

HOTA's novelty lies in formalizing true positive, false positive, and false negative \textit{associations} (links between any two detections to form tracks) as well, and using these to calculate association accuracies that don't depend on matching predicted and ground truth tracks to each other. 

For each true positive pair of detections \textit{c}, the set of TPAs (true positive associations) consists of those TPs with both the same predicted track ID (prID) and ground truth track ID (gtID) as \textit{c}, or: 

\begin{multline}
    \text{TPA}(c)=\{k\},\\
    k\in\{\text{TP}|\text{prID}(k)=\text{prID}(c)\land\text{gtID}(k)=\text{gtID}(c)\}
\end{multline}

Similarly,

\begin{multline}
    \text{FNA}(c)=\{k\},\\
    k\in\{\text{TP}|\text{prID}(k)\neq\text{prID}(c)\land\text{gtID}(k)=\text{gtID}(c)\}\\
    \cup \{\text{FN}|\text{gtID}(k)=\text{gtID}(c)\}
\end{multline}

\begin{multline}
    \text{FPA}(c)=\{k\},\\
    k\in\{\text{TP}|\text{prID}(k)=\text{prID}(c)\land\text{gtID}(k)\neq\text{gtID}(c)\}\\
    \cup \{\text{FP}|\text{prID}(k)=\text{prID}(c)\}
\end{multline}

Then HOTA is computed as:

\begin{equation}
    \text{HOTA}=\sqrt{\frac{\sum_{c\in\text{TP}} \mathcal{A}(c)}{\text{|TP|}+\text{|FP|}+\text{|FN|}}}
\end{equation}
\begin{equation}
    \mathcal{A}(c)=\frac{|\text{TPA}(c)|}{|\text{TPA}(c)|+|\text{FPA}(c)|+|\text{FNA}(c)|}
\end{equation}

HOTA determines associations with respect to each true positive pair of detections (a correct match between a predicted and ground truth detection). In its original formulation, then, the final HOTA number comes from an \textit{average} over localization thresholds of HOTA calculated at each threshold of the similarity score between a predicted and ground truth detection to count as a true positive. 

For bounding box based methods, this similarity score might be IoU. But since \method predicts point tracks and CFC provides bounding box ground truth tracks, our similarity score is always either 1 or 0: a predicted point is matched to a ground truth bounding box if the point is inside the box and it's assigned to that box via a Hungarian matching which optimizes HOTA. We do not average over any thresholds. For the YOLO+SORT results in this paper, we take the centroid of the YOLO-predicted bounding box as the ``predicted point''. This means our HOTA metric is not directly comparable to HOTA calculated conventionally (based on bounding box IoU) on the same dataset.

\end{document}